\documentclass[10pt,twocolumn,letterpaper]{article}

\usepackage[pagenumbers]{iccv} % To force page numbers, e.g. for an arXiv version

\definecolor{iccvblue}{rgb}{0.21,0.49,0.74}
\usepackage[pagebackref,breaklinks,colorlinks,allcolors=iccvblue]{hyperref}
\usepackage[inkscapelatex=false]{svg}
\usepackage{adjustbox}
\usepackage{tabularx}
\usepackage{multirow}
\usepackage{color}
\usepackage{colortbl}
\usepackage{array}
\usepackage{pifont}

\title{Modeling Variants of Prompts 
for Vision-Language Models}

\author{
Ao Li\thanks{Equal contribution to this work.}\textsuperscript{\rm ~~,1} \quad 
Zongfang Liu\footnotemark[1]\textsuperscript{\rm ~~,2} \quad 
Xinhua Li\textsuperscript{1} \quad 
Jinghui Zhang\textsuperscript{1} \quad
Pengwei Wang\thanks{Corresponding author.}\textsuperscript{\rm ~~,1} \quad 
Hu Wang\footnotemark[2]\textsuperscript{\rm ~~,2} \\
\\
\textsuperscript{1}Shandong University \quad \textsuperscript{2}Mohamed bin Zayed University of Artificial Intelligence
}

\begin{document}
\maketitle
\begin{abstract}
Large pre-trained vision-language models (VLMs) offer a promising approach to leveraging human language for enhancing downstream tasks. However, VLMs such as CLIP face significant limitation: its performance is highly sensitive to prompt template design. Although prompt learning methods can address the sensitivity issue by replacing natural language prompts with learnable ones, they are incomprehensible to humans. Ensuring consistent performance across various prompt templates enables models to adapt seamlessly to diverse phrasings, enhancing their ability to handle downstream tasks without requiring extensive prompt engineering. In this work, we introduce the \textbf{RobustPrompt Benchmark}, a systematic benchmark to evaluate robustness to different prompt templates for VLMs. It includes a dataset with hundreds of carefully designed prompt templates, divided into six types, covering a wide variety of commonly used templates. Beside the benchmark, we propose \textbf{M}odeling \textbf{V}ariants of \textbf{P}rompts (MVP), a simple yet effective method that mitigates sensitivity by modeling variants of prompt structures. The innovation of MVP lies in decoupling prompts into templates and class names, and using Variational Autoencoders (VAE) to model the distribution of diverse prompt structures. Experiments across 11 datasets demonstrate that MVP can greatly enhance model robustness to variations in input prompts without a drop in performance. The code is available at \url{https://github.com/liaolea/MVP}.
\end{abstract}

\section{Introduction}
\label{sec:intro}
Recent advances in vision-language models (VLMs) 
like CLIP \cite{radford2021learning} have opened promising new avenues for enhancing downstream tasks through natural language integration. 
VLMs align image and text embeddings through contrastive learning \cite{chen2020simple},
This cross-modal alignment enhances performance on downstream tasks by leveraging task-specific textual information, such as category descriptions, in combination with pre-trained VLMs.

\begin{figure*}[!ht]
    \centering
    \includegraphics[width=1\linewidth]{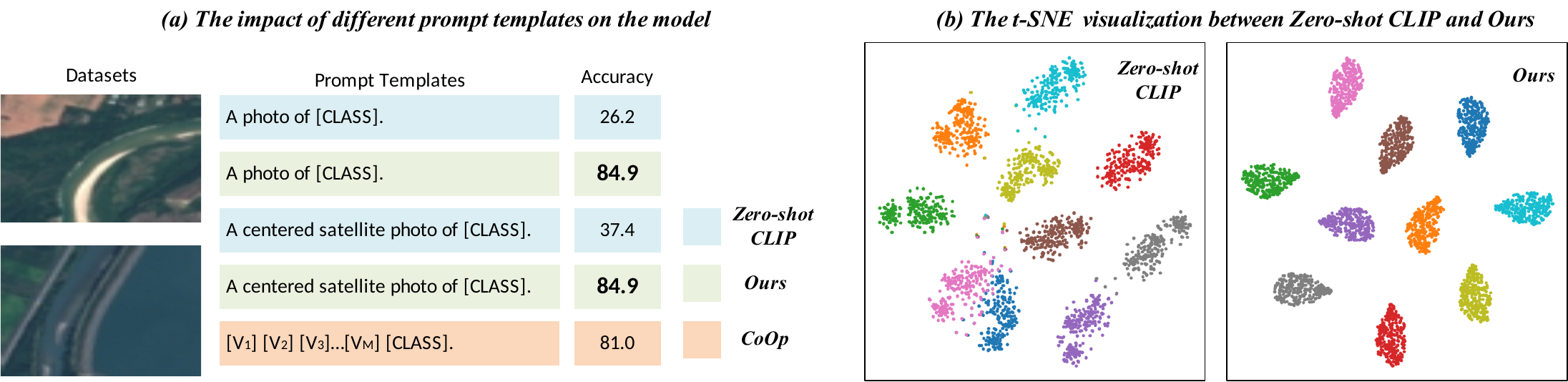}
    \caption{(a) Robustness comparison of zero-shot CLIP and our method on the  EuroSAT dataset. The performance of zero-shot CLIP varies significantly with changes in the prompt template. In contrast, our method maintains strong robustness across different prompt templates. (b) T-SNE visualization of text features. We conduct a T-SNE analysis using 10 classes from the EuroSAT dataset and 300 different prompt templates to visualize the embeddings in the feature space. The text features generated by our method exhibit larger inter-class distances and reduced intra-class variation.} 
    \label{fig:motivation}
\end{figure*}

CLIP is trained on image-text pairs rather than just class names, meaning that it learns to associate images with more descriptive textual information. This is why zero-shot CLIP often relies on prompt templates like ``\texttt{a photo of a \{\}.}" instead of merely using a class name in image classification tasks. However, CLIP faces a significant challenge: its performance is highly sensitive to the structure of prompt templates \cite{zhou2022learning}. For example, ``\texttt{a photo of forest.}" and ``\texttt{a centered satellite photo of forest.}" both convey the same core concept—\textit{forest}—despite their structural differences. However, these variations can result in significant performance differences, as shown in Figure \ref{fig:motivation}(a). Designing effective prompt templates, however, is challenging. In prompt-based multi-modal learning tasks, minor variations—such as changes in tense or articles—can significantly impact model predictions, posing a major challenge. Moreover, users often lack the expertise to craft perfectly tailored prompts. Therefore, enhancing robustness to prompt template variations is crucial.

To bridge this gap, prompt learning methods \cite{derakhshani2023bayesian, zhou2022conditional, zhou2022learning, lu2022prompt}, such as CoOp \cite{zhou2022learning}, have significantly enhanced task-specific performance by fine-tuning learnable tokens. These methods bypass the need for natural language prompting. However, they face a challenge in terms of human interpretability, as the learned prompts are total black boxes to humans, making it difficult to debug, prone to overfitting and have strong model dependencies \cite{zhou2022conditional}. From a completely different perspective, rather than relying on human-unreadable learned prompts, can we enhance the VLMs robustness to diverse natural language prompts while preserving its generality?

Building on this insight and the situation of lacking effective methods to evaluate model robustness, we are the first to propose the \textbf{RobustPrompt Benchmark}, designed to comprehensively evaluate and quantify the model robustness across a range of prompt templates. Through a comprehensive analysis of how different templates affect CLIP's performance, we propose a taxonomy that classifies prompt variations into six types based on natural language attributes and structural differences. We further carefully design templates for each type, forming a comprehensive prompt template benchmark. It provides valuable resources and insights for the community to improve prompt engineering and enhance the model robustness. Then, we propose \textbf{M}odeling \textbf{V}ariants of \textbf{P}rompts (MVP), a method that decouples prompts into templates and class names and models the distribution of various prompt structures using variational autoencoder (VAE)~\cite{kingma2022autoencodingvariationalbayes}. MVP learns a continuous variation space for prompt templates, allowing the model to generate more general and robust representations. As shown in Figure~\ref{fig:motivation}(b), zero-shot CLIP generates text features with large intra-class distances and small inter-class distances in the feature space. In contrast, MVP reduces intra-class variation, demonstrating strong robustness to different templates. It also maintains large inter-class distances, ensuring better class separability and thereby improving performance in few-shot image classification tasks.

In summary, our main contributions are as follows:
\begin{itemize}
\item We first introduce the \textbf{RobustPrompt Benchmark}, a new benchmark that systematically evaluates VLMs robustness to different prompt structures. It quantifies the model’s capacity to handle prompt template variations and offers valuable insights for improving prompt engineering. \par
\item We propose \textbf{MVP}, a simple yet effective approach that mitigates sensitivity by naturally decoupling prompts into templates and class names while leveraging VAE to learn the distribution of diverse prompt structures. \par
\item We conduct extensive experiments on 11 representative datasets, demonstrating that our proposed MVP can greatly enhances model robustness towards the variations of inputted prompt templates without performance drop. \par
\end{itemize}

\section{Related Work}  
\label{sec:related}  
\noindent{\textbf{Vision-Language Models.}}  
Vision-language pre-training has emerged as a promising approach for building models that effectively transfer knowledge across diverse tasks by bridging image content and textual descriptions. Early methods predominantly focused on predicting image captions \cite{joulin2016learning,desai2021virtex,zhang2022contrastive}, but their reliance on relatively small, curated datasets constrained their scalability and generalization capabilities. 

Notable models like CLIP \cite{radford2021learning} and ALIGN \cite{jia2021scaling} use contrastive learning to align image-text representations, pulling matched pairs closer while pushing mismatched pairs apart in the embedding space. This paradigm, driven by natural language supervision, enables VLMs to learn robust visual representations that transfer effectively to downstream tasks. However, while these models excel at generalizing across tasks, their performance remains highly sensitive to the structure and quality of input prompts, limiting their robustness and adaptability in real-world scenarios. Addressing this sensitivity is crucial for unlocking the full potential of VLMs in diverse applications.  

\noindent{\textbf{Prompt Engineering and Prompt Learning.}} Prompt engineering and learning emerged as critical techniques in natural language processing (NLP)~\cite{shin2020autoprompt,brown2020language,li2021prefix}.
Traditional approaches rely on manually designed text prompts~\cite{brown2020languagemodelsfewshotlearners}, while recent research has shifted toward two main directions: first, automating the generation of discrete prompts (i.e., prompts composed of natural language tokens)~\cite{zhou2023largelanguagemodelshumanlevel}; and second, developing learnable ``soft prompt" techniques such as prompt tuning~\cite{lester2021powerscaleparameterefficientprompt}, which optimizes parameters in continuous vector spaces to create prompt representations suitable for various downstream tasks, eliminating the need for labor-intensive handcrafted prompts.

In few-shot learning for classification, CLIP-Adapter \cite{gao2021clip} adds an MLP after the image encoder to get new image features and make residual join with the original image features, which is used to improve the classification performance. Tip-Adapter \cite{zhang2021tip} uses a cache model to store the image features of the few-shot and the corresponding one-hot label in the cache model to assist the zero-shot CLIP to do the classification. It allows for both training-free and fine-tuning. CaFo \cite{zhang2023prompt} generates new image and text features with the help of DINO \cite{zhang2022dinodetrimproveddenoising}, DALL-E \cite{ramesh2021zeroshottexttoimagegeneration}, and GPT-3 \cite{brown2020languagemodelsfewshotlearners}, respectively, greatly improving the classification performance. Though better performance, these methods rely on natural language and are prone to prompt sensitivity.

Methods, e.g., CoOp \cite{zhou2022learning}, extend prompt tuning by learning a prompt to minimize classification loss, effectively replacing human-designed prompts. By using learnable tokens instead of natural language prompts, CoOp addresses the issue of prompt sensitivity. \cite{lu2022prompt, chen2023plotpromptlearningoptimal} further enhance performance. Moreover, \cite{zhou2022conditional, derakhshani2023bayesian, wang2023improving, lee2023readonlypromptoptimizationvisionlanguage} specifically improve performance on unseen classes, thereby strengthening generalization to novel categories.

Since CoOp focuses on fine-tuning human-unreadable tokens, it inevitably faces issues such as being difficult to debug, prone to overfitting, and having strong model dependencies \cite{zhou2022conditional}. Thinking from a totally different perspective, instead of relying on human-incomprehensible tokens, we focus on modeling the variants of natural language prompts, ensuring that the prompts remain human-readable while reducing sensitivity to variations in the input.

\begin{figure*}[ht]
    \centering
    \includegraphics[width=1\linewidth]{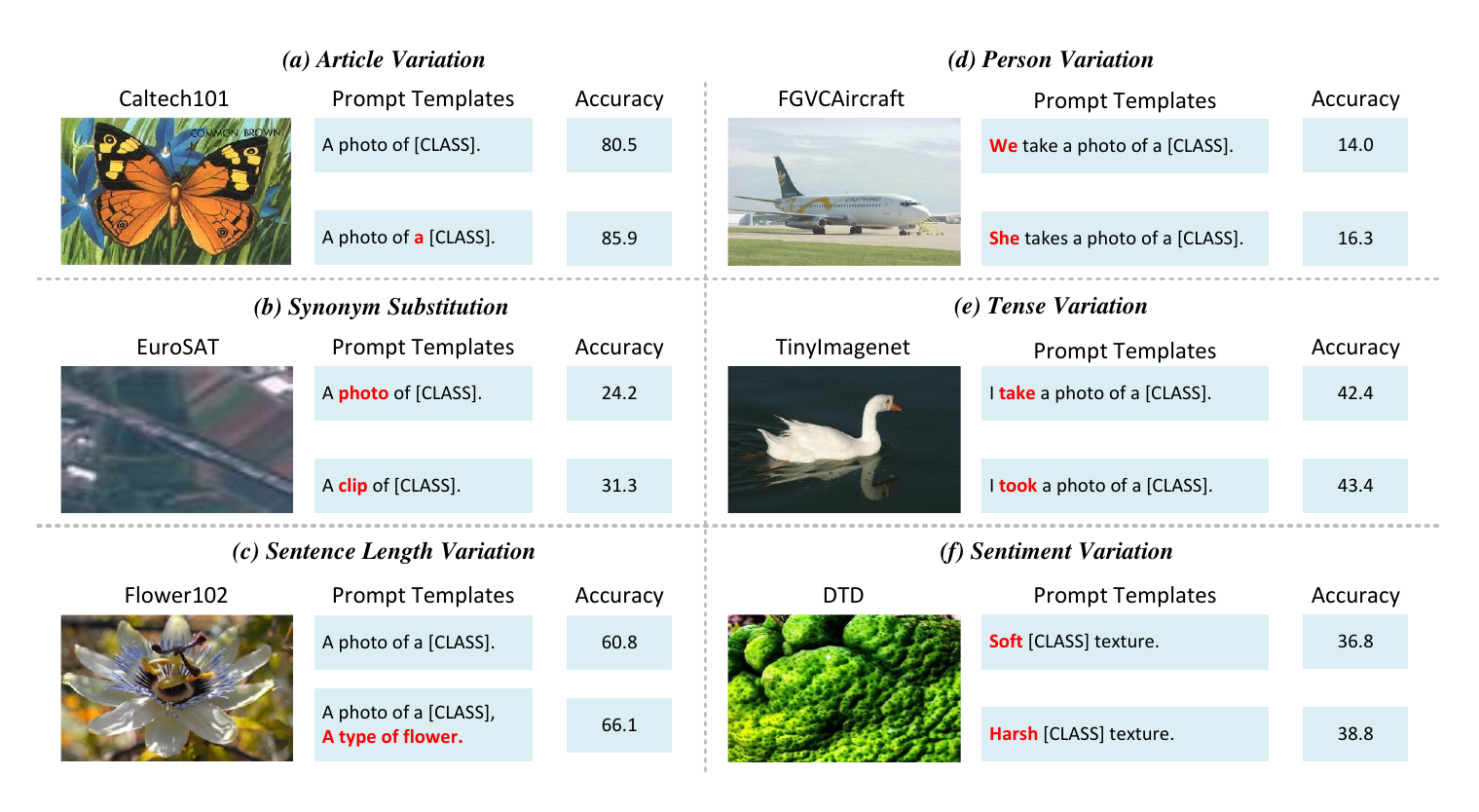}
    \caption{Impact of different prompt templates on zero-shot CLIP. This figure illustrates the performance variations across different prompt template types, highlighting their influence on model robustness.} 
    \label{fig:benchmark}
\end{figure*}

\section{RobustPrompt Benchmark}
\label{sec:benchmark}

\subsection{Prompt Robustness Analysis}
Building on the linguistic properties and structures of prompt templates \cite{justeson1995technical}, we propose a taxonomy to identify and categorize six types of template variations. We further divide each type into subtypes, with each subtype comprising one or more semantically related templates:

\noindent{\ding{182}~\textbf{Article Variation.}}
The presence or absence of an article, such as ``a" or ``the", impacts CLIP’s performance, as shown in Figure~\ref{fig:benchmark}(a). It can influence how the model interprets the described category, making the description either more specific or more ambiguous. This type consists of two subtypes: one with an article and one without. For instance, ``\texttt{a photo of a flower.}" refers to a specific instance of a flower, whereas ``\texttt{a photo of flower.}" conveys a more general description.

\noindent{\ding{183}~\textbf{Synonym Substitution.}}
Synonym substitution influences CLIP’s performance, as replacing ``photo" with alternatives such as ``image" or ``frame" often impacts its behavior. This type consists of two subtypes: using the common word ``photo" and using alternative words.  As shown in Figure~\ref{fig:benchmark}(b), in remote sensing imagery, the term   ``clip" may be more appropriate than ``photo" to describe the segment, potentially leading to better performance.

\noindent{\ding{184}~\textbf{Sentence Length Variation.}} Sentence length variation refers to changing the length of a prompt by adding or removing details. This type consists of two subtypes: short sentences and long sentences. A short template like ``\texttt{a photo of \{\}.}" is more general, whereas a longer version, such as ``\texttt{a photo of \{\}, a type of flower.}" introduces more information. In some cases, adding details can increase ambiguity and affect performance, while in other cases, as shown in Figure~\ref{fig:benchmark}(c), the additional context can clarify the category, aiding the model’s interpretation.

\noindent{\ding{185}~\textbf{Person Variation.}}
The selection of grammatical person in pronoun usage introduces subtle perceptual nuances in the interpretation of actions.
This type encompasses three subtypes: first-person, second-person, and third-person perspectives. 
For example, as shown in Figure~\ref{fig:benchmark}(d), the descriptions   ``\texttt{We take a photo of a flower.}" (first-person) and ``\texttt{She takes a photo of a flower.}" (third-person) convey identical scenarios but from different perspectives, potentially affecting the model's ability to align the text with the image.

\noindent{\ding{186}~\textbf{Tense Variation.}}
Tense variation, which involves altering the verb tense in a prompt, can affect performance. This type consists of three subtypes: present tense, past tense, and future tense. For example, as shown in Figure~\ref{fig:benchmark}(e), changing ``\texttt{takes a photo}" (present tense) to ``\texttt{took a photo}" (past tense) or ``\texttt{will take a photo}" (future tense) alters the temporal aspect of the prompt, which in turn can influence performance.

\noindent{\ding{187}~\textbf{Sentiment Variation.}}
Sentiment variation involves modifying the emotional tone of the prompt. This type consists of two subtypes: positive and negative. For example, as shown in Figure~\ref{fig:benchmark}(f), changing a template like ``\texttt{Soft \{\} texture.}"(positive sentiment) to ``\texttt{Harsh \{\} texture.}" (negative sentiment) can shift the sentiment expressed.

We adopt GPT-4 API\footnote{\url{https://openai.com/index/gpt-4/}} to generate a diverse set of templates for each type, resulting in 1,000 distinct prompt templates. All templates are used for evaluation to analyze the performance variation of zero-shot CLIP across different template types in multiple datasets. Figure~\ref{fig:benchmark} presents randomly selected examples for each type, demonstrating intra-type performance fluctuations in the six template types. These findings underscore the significant influence of template type on assessing model robustness.

\subsection{Robust Prompt Dataset}
Due to the current lack of datasets for prompt robustness evaluation, we create a dataset by refining the previously mentioned 1,000 distinct templates. To ensure that the selected templates comprehensively represent common linguistic variations, we conduct rigorous manual filtering within each type. To optimize the template set, we combine the `person variation' and `tense variation' types, as well as merge `article variation' with `synonym substitution' to form more cohesive types. Additionally, we develop an extended set of templates that provide more detailed and specific descriptions. This process ultimately leads to the creation of our robust prompt dataset. 

We divide the dataset into non-overlapping training and testing sets for each template type. Details of the dataset are provided in the supplementary materials.

\begin{table}[ht]
    \centering
    \small
    \begin{adjustbox}{max width=\linewidth}
    \begin{tabular}{lr}
        \toprule
        \textbf{Types} & \textbf{Number of templates} \\ 
        \midrule
        Article and Synonym Variation & 155 \\ 
        Sentence Length Variation & 86 \\ 
        Sentiment Variation & 72 \\ 
        Person and Tense Variation & 24 \\ 
        Detailed Description & 396 \\ 
        \midrule
        \textbf{Total} & \textbf{733} \\ 
        \bottomrule
    \end{tabular}
    \end{adjustbox}
    \caption{Distribution of prompt templates across the five consolidated variation types in the robust prompt dataset.}
    \label{tab:prompt_dataset}
\end{table}

\begin{figure*}[ht]
    \centering
    \includegraphics[width=1\linewidth]{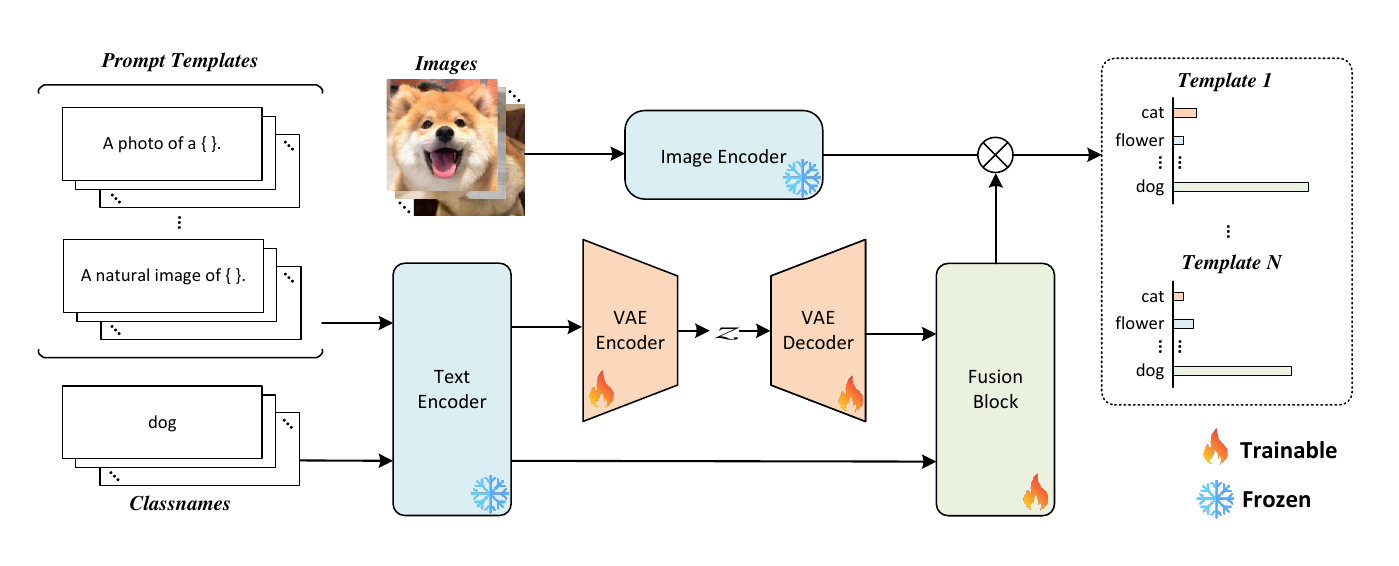}
    \caption{The MVP Architecture.
    MVP primarily consists of CLIP, VAE, and a fusion block. 
    (1) During each training epoch, we randomly select \textit{a subset of templates} from the robust prompt dataset. We naturally decouple the original prompt into two components: prompt templates and class names, which are then processed separately by the CLIP text encoder. 
    The template component is processed by the VAE to model the distribution of templates, while the class names are directly encoded. 
    Subsequently, features from both components are concatenated and integrated through the fusion block. Finally, the resulting fused representation is used to compute cosine similarity with image features for classification. 
    (2) During inference, only \textit{a single prompt template} is provided as input.} 
    \label{fig:model}
\end{figure*}

\subsection{Prompt Robustness Score}
When a model demonstrates consistent performance across subtypes (e.g., showing identical performance on both positive and negative sentiment templates), it indicates robustness to variations within that type. To quantitatively evaluate this robustness, we introduce the \textbf{Prompt Robustness Score (PRS)}, which measures the relative performance gap among template subtypes.
For each type, we identify the best-performing subtype and calculate the difference between its performance and the average performance across all other subtypes within the same type. This difference is then normalized by the performance of the best-performing subtype. 
Specifically, let \( S_i \) represent the performance of the \(i\)-th subclass, where \( i \in \{1, 2, \dots, n\} \) denotes the set of all subtypes for a given type. 
We assume that \( i^{\prime} \) corresponds to the best-performing subclass, i.e., \( S_{i^{\prime}} = \max S_i \), The prompt robustness score is then defined as:

\begin{equation}
PRS = \frac{\left| S_{i^{\prime}} - \frac{1}{n-1} \sum_{j \neq i^{\prime}} S_j \right|}{S_{i^{\prime}}} \times 100\%,
\end{equation}

A lower PRS value indicates greater robustness to prompt template variations, as it reflects smaller performance disparities across different templates.

\section{Methodology}

\begin{table*}[t]\small
    \centering
    \vspace{-4mm}
    \addtolength\tabcolsep{-2.4pt} 
    \begin{adjustbox}{max width=\textwidth}
    % \resizebox{1.0\linewidth}{!}{
    \begin{tabular}{>{\raggedright\arraybackslash}m{1.5cm}>{\raggedleft\arraybackslash}m{1.3cm}>{\raggedleft\arraybackslash}m{1.3cm}>{\raggedleft\arraybackslash}m{1.3cm}>{\raggedleft\arraybackslash}m{1.3cm}>{\raggedleft\arraybackslash}m{1.3cm}>{\raggedleft\arraybackslash}m{1.3cm}>{\raggedleft\arraybackslash}m{1.3cm}>{\raggedleft\arraybackslash}m{1.3cm}>{\raggedleft\arraybackslash}m{1.3cm}>{\raggedleft\arraybackslash}m{1.3cm}>{\raggedleft\arraybackslash}m{1.3cm}}%{lccccccccccc}
    \toprule

    \textbf{Datasets} & ImageNet & OxfordPets & Flowers102 & FGVCAircraft & DTD & EuroSAT & StanfordCars & Food101 & SUN397 & Caltech101 & UCF101 \\

    \midrule
    \rowcolor{gray!15}\multicolumn{12}{c}{\textbf{Zero-shot CLIP}} \\
    PRS-Type1 & 4.144 & 3.861 & 2.912 & 1.021 & 3.550 & 6.797 & 1.004  & 0.947 & 0.858 & 6.170 & 1.194 \\
    PRS-Type2 & 4.159 & 5.088 & 5.936 & 5.999 & 5.079 & 22.684 & 2.855 & 7.451 & 6.874 & 3.463 & 2.128 \\
    PRS-Type3 & 0.975 & 2.494 & 7.866 & 8.842 & 9.862 & 25.196 & 0.652 & 3.586 & 3.111 & 0.486 & 6.058 \\
    PRS-Type4 & 2.730 & 4.620 & 6.019 & 9.053 & 2.540 & 8.160  & 7.630 & 2.521 & 2.053 & 2.206 & 1.983 \\
    PRS-Type5 & 0.144 & 0.667 & 2.468 & 0.831 & 0.246 & 17.634 & 3.323 & 0.664 & 0.078 & 1.951 & 3.757 \\
    PRS-Type6 & 1.368 & 1.302 & 1.614 & 2.058 & 5.847 & 0.902  & 1.566 & 0.310 & 1.322 & 1.658 & 0.170 \\
    \midrule
    PRS-Avg & 2.253 & 3.005 & 4.469 & 4.634 & 4.521 & 13.562 & 2.838 & 2.580 & 2.383 & 2.656 & 2.548 \\

    \midrule
    \rowcolor{gray!15}\multicolumn{12}{c}{\textbf{CLIP-Adapter}} \\ 
    PRS-Type1 & 7.065 & 5.490 & 5.610 & 9.957 & 1.959 & 0.988 & 0.407 & 0.260 & 3.406 & 2.660 & 0.181 \\
    PRS-Type2 & 2.827 & 2.033 & 1.163 & 4.069 & 2.332 & 1.345 & 3.042 & 1.569 & 2.880 & 0.443 & 0.434 \\
    PRS-Type3 & 1.567 & 0.034 & 1.686 & 3.910 & 1.776 & 4.876 & 0.413 & 1.500 & 0.395 & 0.443 & 3.595 \\
    PRS-Type4 & 2.904 & 4.873 & 1.556 & 3.081 & 0.972 & 0.929 & 3.472 & 2.260 & 1.270 & 0.670 & 0.663 \\
    PRS-Type5 & 2.591 & 0.812 & 0.606 & 4.483 & 1.201 & 1.428 & 2.617 & 1.556 & 1.458 & 0.708 & 0.352 \\
    PRS-Type6 & 0.227 & 0.283 & 0.072 & 0.130 & 0.381 & 1.617 & 0.992 & 0.272 & 0.598 & 0.294 & 1.088 \\
    \midrule
    PRS-Avg & 2.863 & 2.254 & 1.782 & 4.272 & 1.437 & 1.864 & 1.824 & 1.236 & 1.668 & 0.870 & 1.052 \\

    \midrule
    \rowcolor{gray!15}\multicolumn{12}{c}{\textbf{Tip-Adapter-F}} \\
    PRS-Type1 & 4.823 & 5.851 & 4.903 & 4.742 & 0.105 & 0.930 & 0.125 & 0.771 & 3.005 & 1.240 & 0.036 \\
    PRS-Type2 & 3.929 & 2.051 & 3.689 & 8.904 & 2.103 & 4.155 & 1.351 & 5.582 & 3.212 & 0.709 & 0.756 \\
    PRS-Type3 & 2.224 & 1.425 & 4.275 & 7.452 & 9.265 & 11.034 & 0.136 & 3.995 & 0.809 & 0.044 & 2.833 \\
    PRS-Type4 & 4.885 & 6.587 & 1.423 & 4.251 & 1.775 & 1.324 & 2.385 & 5.200 & 1.114 & 0.606 & 0.673 \\
    PRS-Type5 & 4.143 & 2.731 & 1.118 & 1.135 & 0.211 & 2.870 & 2.117 & 3.745 & 1.952 & 0.975 & 0.258 \\
    PRS-Type6 & 0.873 & 0.350 & 0.218 & 0.509 & 2.527 & 2.667 & 0.167 & 0.407 & 1.118 & 0.317 & 0.307 \\
    \midrule
    PRS-Avg & 3.480 & 3.166 & 2.604 & 4.499 & 2.664 & 3.830 & 1.047 & 3.283 & 1.868 & 0.649 & 0.810 \\

    \midrule
    \rowcolor{yellow!50}\multicolumn{12}{c}{\textbf{Ours (MVP)}} \\
    PRS-Type1 & 0.012 & 0.000   & 0.000   & 0.109 & 0.094 & 0.126 & 0.067 & 0.025 & 0.003 & 0.015 & 0.048 \\
    PRS-Type2 & 0.014 & 0.158 & 0.014 & 0.081 & 0.094 & 0.049 & 0.079 & 0.002 & 0.025 & 0.044 & 0.060 \\
    PRS-Type3 & 0.014 & 0.124 & 0.028 & 0.027 & 0.094 & 0.165 & 0.062 & 0.016 & 0.003 & 0.030 & 0.095 \\
    PRS-Type4 & 0.014 & 0.096 & 0.011 & 0.136 & 0.026 & 0.060 & 0.022 & 0.017 & 0.013 & 0.047 & 0.089 \\
    PRS-Type5 & 0.005 & 0.023 & 0.000 & 0.163 & 0.031 & 0.131 & 0.011 & 0.033 & 0.041 & 0.000   & 0.107 \\
    PRS-Type6 & 0.011 & 0.011 & 0.014 & 0.095 & 0.037 & 0.034 & 0.002 & 0.003 & 0.002 & 0.012 & 0.054 \\
    \midrule
    PRS-Avg & \textbf{0.012} & \textbf{0.069} & \textbf{0.011} & \textbf{0.102} & \textbf{0.063} & \textbf{0.094} & \textbf{0.040} & \textbf{0.016} & \textbf{0.014} & \textbf{0.025} & \textbf{0.076} \\

    \bottomrule
    \end{tabular} %}
    \end{adjustbox}
    \vspace{-2mm}
    \caption{Main results of PRS on 11 datasets. The values in the table omit \%. MVP significantly outperforms other models, achieving excellent performance in terms of prompt robustness.}
    \vspace{-2mm}
    \label{tab:robustness} 
\end{table*}

In this section, we introduce MVP, the novel approach we propose for robustly modeling prompt variants. The innovation of MVP lies in decoupling prompts into templates and class names, and utilizing a VAE to model the distribution of diverse prompt structures. The framework of MVP is illustrated in Figure \ref{fig:model}.

\subsection{Modeling the Distribution of Diverse Prompt Structures}
Even a minor difference between templates, such as ``\texttt{a photo of a \{\}.}" and ``\texttt{a photo of \{\}.}", can lead to significant performance variations, as shown in Figure~\ref{fig:benchmark}(a). 
This challenge parallels a fundamental limitation in autoencoder (AE), where the model struggles to map similar samples to adjacent locations in the latent space. This limitation stems from the discrete nature of AE's latent representations, as opposed to a continuous probabilistic distribution.
In contrast, VAE introduce probabilistic constraints, making the latent space smoother and more structured, effectively addressing the issue. We hypothesize that CLIP exhibits behavior similar to that of AE. Based on this assumption, we aim to leverage VAE to model the distribution of prompt structures.

Modeling the distribution of prompt structures requires a diverse set of templates. Furthermore, we observe that models trained with a single template tend to overfit that specific template, leading to poor robustness when exposed to other templates (Please refer to section 3 of the supplementary materials). 
To mitigate these issues, we use templates from the robust prompt dataset as our data source, as the diverse templates in this dataset effectively capture a wide range of variations. Instead of relying on a fixed template, we adopt a dynamic training strategy in which, during each epoch, a subset of templates is randomly sampled from the robust prompt dataset. This approach enables the model to learn from diverse template features sampled by the VAE throughout training, thereby enhancing its adaptability to various templates.

Intuitively, concatenating the template with the class name to form a prompt while still expecting the VAE to model the template distribution is unreasonable. For example, ``\texttt{a photo of a cat.}" and ``\texttt{a photo of a dog.}" should be modeled into one prompt distribution as VAE is not responsible for modeling classification but rather the prompt distributions.
To address this issue, we propose a strategy called \textbf{template-class decoupling}, where the prompt is fully decoupled into two components: the template and the class name. The class name is directly processed by the CLIP text encoder, while the template is first passed through the text encoder and then reconstructed by a VAE, ensuring that both components are accurately represented. Finally, the two components are fused using a fusion block, enabling the model to simultaneously learn the robustness of the template and the semantics of the class name.

Specifically, MVP consists of an image encoder \( f \), a text encoder \( g \), a VAE \( v \), and a fusion block \( h \). Given a downstream image classification dataset with \( K \) classes, the set of class names is defined as \( C = \{ c_1, c_2, \dots, c_K \} \). For an input image \( I \), the extracted image features are \( f(I) \). In each training epoch, we randomly sample \( M \) templates from the robust prompt dataset, denoted as \( T = \{ t_1, t_2, \dots, t_M \} \). We adopt the template-class decoupling strategy, where the class names and templates are separately encoded using the text encoder \( g \). This results in class name features \( g(C) \) and template features \( g(T) \), respectively, where \(g(C) = \{ g(c_1), g(c_2), \dots, g(c_K) \}\)
and \(g(T) = \{ g(t_1), g(t_2), \dots, g(t_M) \}.\)
We use VAE to reconstruct the template features, and concatenate the reconstructed template features with the class name features and feed the combined representation into the fusion block for integration. Finally, we compute the dot product with the image features to obtain the final logits:
 
\begin{equation}
     y_{ij} = f(I) \cdot h (\text{cat}(v(g(t_{i})), g(c_{j}))),
\end{equation}
where \(y_{ij}\) represents the logits of the \(i\)-th template and the \(j\)-th class name with the image \(I\).

\begin{figure*}[ht]
    \centering
    \includegraphics[width=1\linewidth]{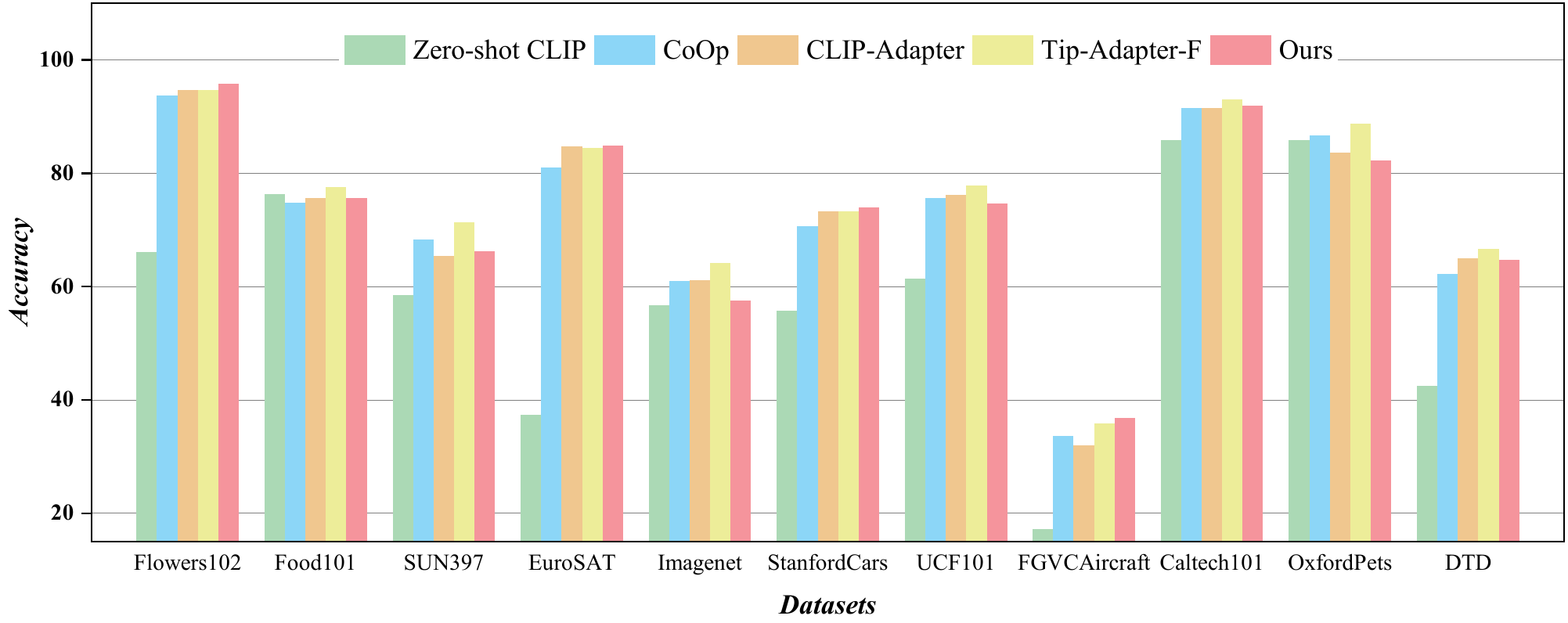}
    \caption{Main results of accuracy on 11 datasets. MVP outperforms zero-shot CLIP, CoOp, and CLIP-Adapter in overall accuracy, while demonstrating competitive performance with Tip-Adapter-F, showing strengths and weaknesses across different datasets.}
    \label{fig:acc}
\end{figure*}

\subsection{Multi-template Loss and VAE Loss}
Throughout the training process, we incorporate two essential loss functions: the multi-template loss and the VAE loss. The multi-template loss computes the classification loss by applying the cross-entropy loss function to each template independently, enabling the model to learn the classification task from each template separately. In parallel, we employ the VAE loss to guide the reconstruction of the template features, allowing the model to capture the latent structure of the prompts.

Specifically, the multi-template loss is defined as:
\begin{equation}\small
\mathcal{L}_{\text{MT}} = -\sum_{i=1}^{M}\sum_{j=1}^{K} y^{I} logy_{ij},
\end{equation}
where \( y^{I} \) represents the ground truth label for the image \(I\). The VAE loss is computed as follows:

\begin{equation}\small
\mathcal{L}_{\text{VAE}} =  \sum_{i=1}^{M}(\| g(t_i) - v(g(t_i)) \|_2^2 + D_{\text{KL}}(q(z | t_{i}) \parallel p(z))), 
\end{equation}
where \( z \) is the latent variable sampled from the approximate posterior distribution \( q(z | t_i) \), and \( D_{\text{KL}} \) denotes the Kullback-Leibler (KL) divergence. \( p(z)\) represents the prior distribution, typically assumed to be a standard Gaussian distribution $\mathcal{N}(0, \mathbf{I})$. 

Hence, the overall loss is computed as follows:

\begin{equation}\small
\mathcal{L}_{\text{All}} = \mathcal{L}_{\text{MT}} + \alpha\mathcal{L}_{\text{VAE}},
\end{equation}where \(\alpha\) is a loss coefficient.

\section{Experiment}  
\label{sec:experiment} 

\subsection{Experimental Setup}
We select 11 publicly available image classification datasets, following the setup of CoOp: ImageNet \cite{5206848}, Caltech101 \cite{fei2004learning}, OxfordPets \cite{parkhi2012cats}, StanfordCars \cite{krause20133d}, Flower102 \cite{nilsback2008automated}, Food101 \cite{bossard2014food}, FGVCAircraft \cite{maji2013fine}, SUN397 \cite{xiao2010sun}, DTD \cite{cimpoi2014describing}, EuroSAT \cite{helber2019eurosat}, and UCF101 \cite{soomro2012ucf101}.

We compare our approach with several existing methods in terms of prompt robustness and classification accuracy. For prompt robustness, we evaluate methods that rely solely on text prompts, including zero-shot CLIP, CLIP-Adapter, and Tip-Adapter-F. For classification accuracy, we compare our method with zero-shot CLIP, CoOp, CLIP-Adapter, and Tip-Adapter-F. All models, except for CoOp,  are evaluated using the templates designed by CLIP.

\subsection{Implementation details}
We train MVP using AdamW~\cite{loshchilov2017decoupled} with a learning rate of 0.001, applying a warm-up cosine annealing strategy. 
The batch size is set to 32, and 50 templates are randomly sampled for each epoch. 
For few-shot learning, the number of shots is set to 16. 
MVP utilizes ResNet-50 \cite{he2015deepresiduallearningimage} as visual backbone and BERT \cite{devlin2019bertpretrainingdeepbidirectional} as the text backbone. We set the VAE loss weight \( \alpha\) to 1, with a latent code dimension of 128. The VAE encoder consists of two fully connected layers, while both the fusion block and the VAE decoder are implemented as single linear layers with GELU activation. All experiments are conducted on a single NVIDIA RTX 3090 GPU.

\subsection{Results}
\noindent{\textbf{Prompt Robustness.}}
PRS-Type1 to PRS-Type6 correspond one-to-one to the six variations in Section \ref{sec:benchmark}. As illustrated in Table \ref{tab:robustness}, other methods exhibit varying degrees of sensitivity across different datasets and test types. For instance, in PRS-Type 3, zero-shot CLIP performs better on the EuroSAT dataset with longer, more informative prompts than with shorter, less descriptive ones. Similarly, in PRS-Type 1, CLIP-Adapter exhibits significant sensitivity on the FGVCAircraft dataset, where the template ``\texttt{a photo of a \{\}.}" performs notably better than ``\texttt{a photo of \{\}.}". In contrast, our approach achieves substantial improvements in PRS, with values  approaching zero in nearly all cases, highlighting its superior robustness.

\noindent{\textbf{Few-shot Learning.}} 
As show in Figure \ref{fig:acc}, in terms of classification accuracy for few-shot learning, our approach outperforms zero-shot CLIP, CoOp, and CLIP-Adapter. On different datasets, our method shows a competitive performance, with its strengths and weaknesses varying when compared to Tip-Adapter-F.

\subsection{Analysis}
We select ImageNet and the fine-grained datasets EuroSAT and FGVC Aircraft for analysis.

\begin{figure}[htbp]
    \centering
    \includegraphics[width=0.9\linewidth]{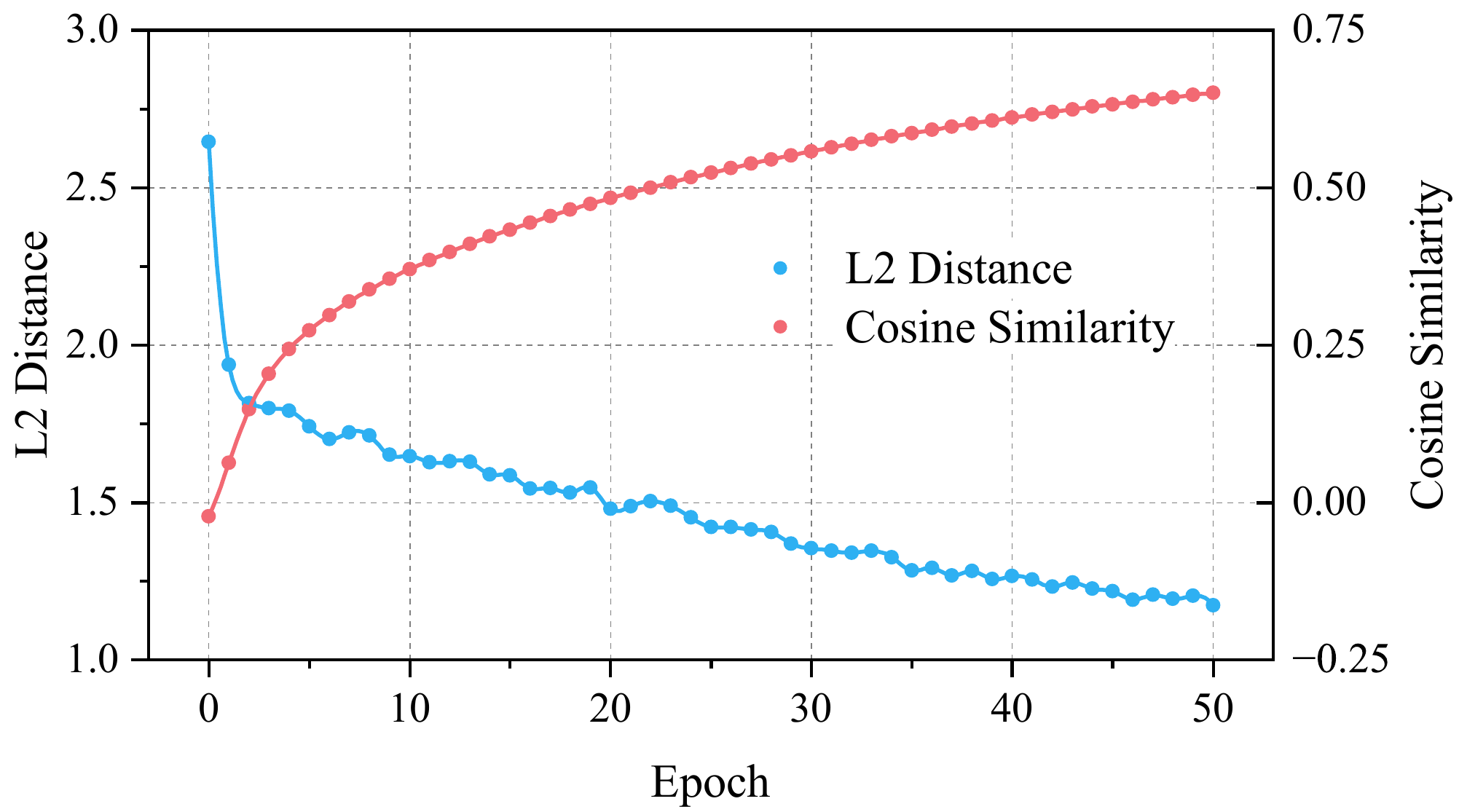}
    \caption{Visualization of the L2 distance and cosine similarity between the text features generated by CLIP text encoder and the text features reconstructed by the VAE on the ImageNet dataset.} 
    \label{fig:l2}
\end{figure}

\noindent{\textbf{VAE Reconstruction of Text Features.}} To verify the effectiveness of VAE reconstruction, we train MVP for 50 epochs on the ImageNet dataset and compute the L2 distance and cosine similarity between the original and reconstructed text features. As shown in Figure~\ref{fig:l2}, the decreasing L2 distance and the increasing cosine similarity approaching 1 demonstrate the effectiveness of the reconstruction.

\noindent{\textbf{Variants with MVP.}} We investigate three variants of MVP:
1):\textbf{w/o decoupling}: Templates and class names are concatenated and then directly passed to both the text encoder, VAE, and fusion block.
2):\textbf{w/o VAE}: Templates and class names are processed separately by the text encoder, concatenated, and then passed into the fusion block.
3):\textbf{w/o decoupling and VAE}: Templates and class names are concatenated and then directly passed to both the text encoder and fusion block.
Additionally, to ensure convergence for the \textbf{w/o decoupling} variant, we apply a scaling factor of 0.1 to the variance during the VAE resampling process and set the VAE loss coefficient \( \alpha \) to 0.01. For the \textbf{w/o VAE} and \textbf{w/o decoupling and VAE} variants, as the model does not use VAE, we omit the VAE loss.

\begin{figure}[htbp]
    \centering
    \includegraphics[width=0.9\linewidth]{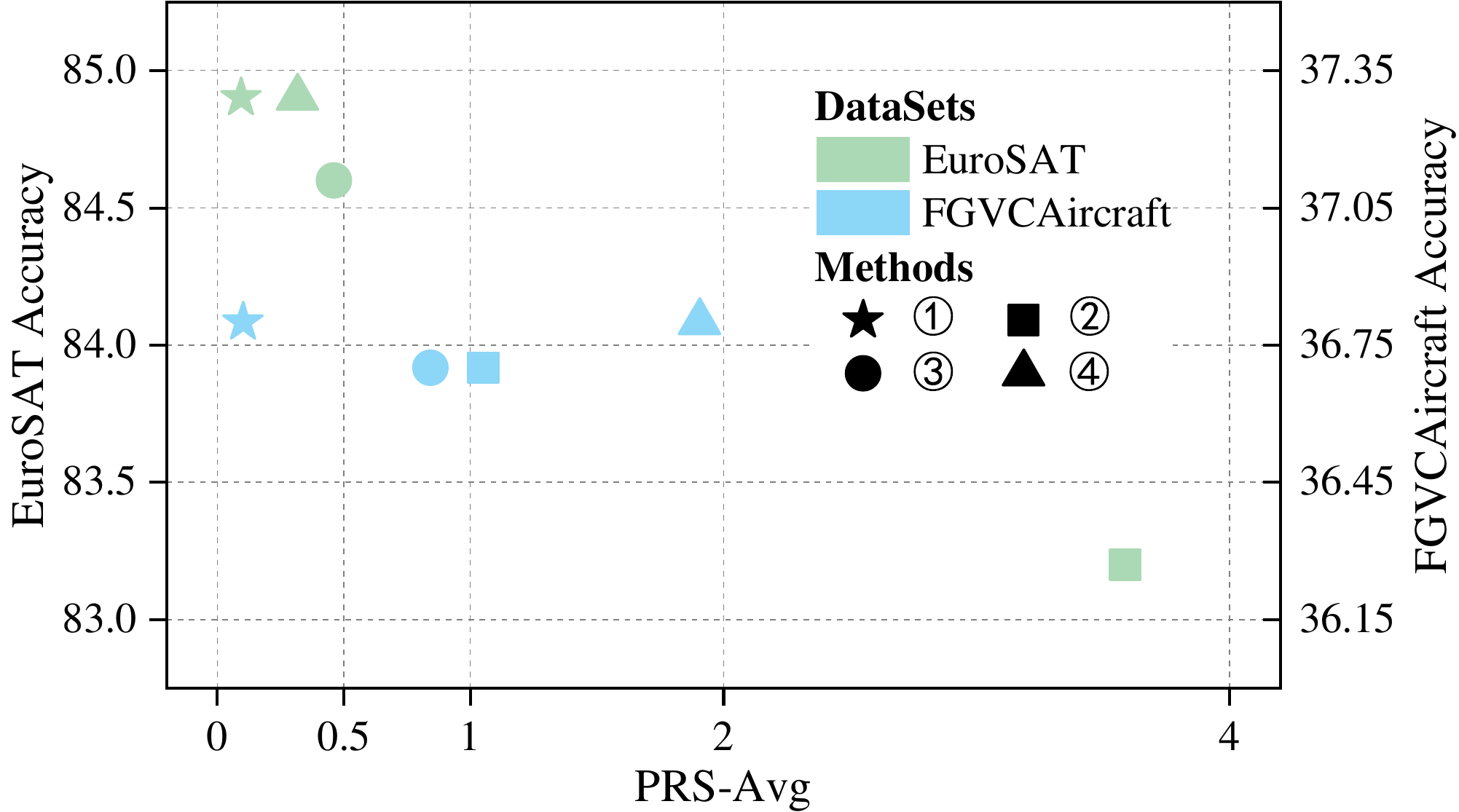}
    \caption{PRS-Avg vs. Accuracy for different variants of MVP (Top-left is better), where the x-axis represents the PRS-Avg value and the y-axis represents accuracy. \ding{172} represents our method, \ding{173} represents w/o decoupling, \ding{174} represents w/o VAE, and \ding{175} represents w/o decoupling and VAE.} 
    \label{fig:ablation1}
\end{figure}

As shown in Figure \ref{fig:ablation1}, all variants, except ours, suffer from prompt sensitivity issues across both datasets. In the \textbf{w/o decoupling} variant, where templates and class names are processed together in the VAE, the model ends up learning a shared representation for both, making it challenging for the VAE to model a distribution that is robust across different templates while still distinguishing between classes. This leads to a drop in accuracy.
The \textbf{w/o VAE} variant achieves better accuracy and robustness than the \textbf{w/o decoupling} variant. However, despite these improvements, it still falls short when compared to ours.
The \textbf{w/o decoupling and VAE} variant shows accuracy comparable to ours, but its robustness still falls short in comparison. 

These results emphasize the importance of template-class decoupling and modeling the distribution of diverse prompt structures.

\noindent{\textbf{The Impact of the Number of Randomly Sampled Templates.}} We investigate the impact of varying the number of templates sampled per epoch. We consider four different sample sizes: 10, 30, 50, and 70.

\begin{figure}[htbp]
    \centering
    \includegraphics[width=0.9\linewidth]{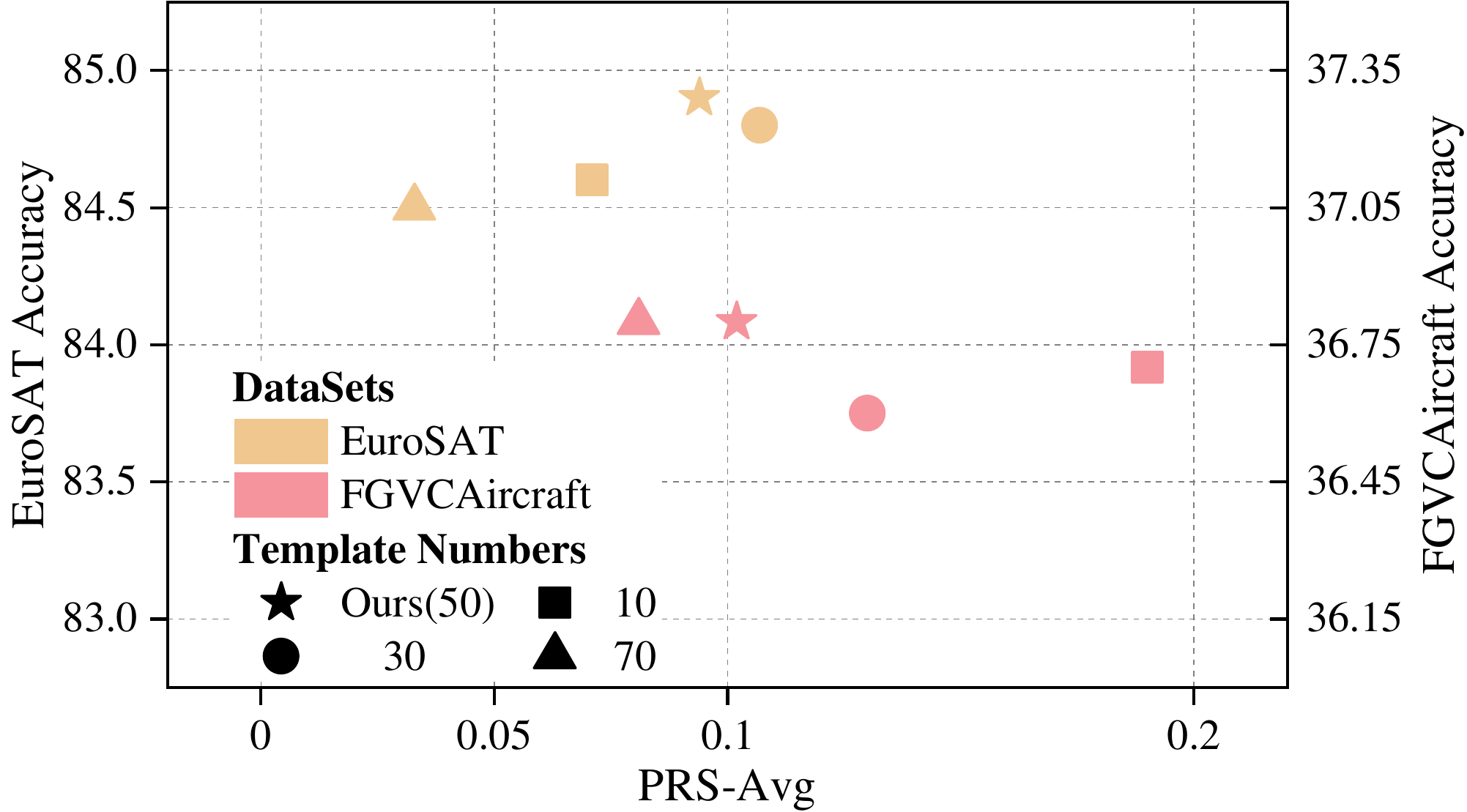}
    \caption{PRS-Avg vs. Accuracy for different template samples per epoch (Top-left is better), where the x-axis represents the PRS-Avg value and the y-axis represents accuracy.} 
    \label{fig:ablation3}
\end{figure}

Intuitively, increasing the number of templates allows the VAE to capture a broader structure of prompt variations. As shown in Figure \ref{fig:ablation3}, initially, as the number of sampled templates increases, both robustness and accuracy do not show significant changes. However, when the sampling number reaches 50, the templates learned by the model in each epoch become more diverse, leading to improvements in both accuracy and robustness. When the sampling number reaches 70, while robustness continues to improve, accuracy does not surpass our model's performance. Overall, 50 templates emerge as the optimal choice for balancing both accuracy and robustness.

\section{Conclusion}
We propose the \textbf{RobustPrompt Benchmark}, categorizing factors affecting prompt robustness and creating the robust prompt dataset to evaluate VLMs robustness against template variations. 
Furthermore, we introduce \textbf{MVP}, which disentangles prompts into templates and class names and employs a VAE to model the distribution of diverse prompt structures. 
Experiments reveal that zero-shot CLIP and other text input-based approaches suffer from prompt sensitivity issues, confirming our hypothesis that CLIP exhibits behavior similar to that of AE. In contrast, MVP effectively mitigates this sensitivity, demonstrating superior prompt robustness while maintaining competitive few-shot classification accuracy. Additionally, ablation studies validate the effectiveness of the model components, further reinforcing the necessity of our design. We hope that this work contributes to future advancements in prompt engineering and model robustness. %

% \newpage

{
    \small
    \bibliographystyle{ieeenat_fullname}
    \bibliography{main}
}

\end{document}